\documentclass[twocolumn, switch]{article} % Method A for two-column formatting

\usepackage{preprint}
\usepackage[dvipsnames, table]{xcolor}
\usepackage{amsmath, amsthm, amssymb, amsfonts}
\usepackage{bm}
\usepackage{amsmath}
\usepackage{graphicx}
\usepackage{enumerate}
\usepackage{floatrow}
\usepackage[algoruled,algo2e]{algorithm2e}

\usepackage{setspace}
\usepackage{amssymb}
\usepackage{booktabs}
\usepackage{array}
\usepackage{color}
\usepackage{multirow}
\usepackage{multicol}
\usepackage{threeparttable}
\usepackage{url}
\usepackage[page]{appendix}

\usepackage[numbers,square]{natbib}
\usepackage[utf8]{inputenc}	% allow utf-8 input
\usepackage[T1]{fontenc}	% use 8-bit T1 fonts
\usepackage[colorlinks = true,
            linkcolor = blue,
            urlcolor  = blue,
            citecolor = blue,
            anchorcolor = black]{hyperref}	% Color links to references, figures, etc.
\usepackage{booktabs} 		% professional-quality tables
\usepackage{nicefrac}		% compact symbols for 1/2, etc.
\usepackage{microtype}		% microtypography
\usepackage{lineno}		% Line numbers
\usepackage{float}			% Allows for figures within multicol
\usepackage{newfloat}
\DeclareFloatingEnvironment[name={Supplementary Figure}]{suppfigure}
\usepackage{sidecap}
\sidecaptionvpos{figure}{c}
\usepackage{floatrow}
\usepackage{graphicx}
\usepackage{subcaption}
\usepackage{amssymb}

\usepackage{algorithm}
\usepackage{algorithmic}
\definecolor{Gray}{gray}{0.93}

\usepackage{titlesec}
\titlespacing\section{0pt}{12pt plus 3pt minus 3pt}{1pt plus 1pt minus 1pt}
\titlespacing\subsection{0pt}{10pt plus 3pt minus 3pt}{1pt plus 1pt minus 1pt}
\titlespacing\subsubsection{0pt}{8pt plus 3pt minus 3pt}{1pt plus 1pt minus 1pt}

\usepackage{graphics}
\usepackage{hyperref}
\usepackage{color}
\usepackage{multirow}
\usepackage{hhline}
\usepackage{lipsum}
\usepackage{pifont}
\usepackage{amsfonts} 
\usepackage{colortbl} 

\usepackage{amsmath}

\usepackage{multirow}
\newcommand{\up}{\textcolor{ForestGreen}{$\uparrow$}}
\newcommand{\dn}{\textcolor{BrickRed}{$\downarrow$}}

\title{LOP: Learning Optimal Pruning for Efficient On-Demand MLLMs Scaling
}
\usepackage{authblk}

\author[1]{Zhihan Zhang}
\author[2]{Xiang Pan}
\author[1]{Hongchen Wei}
\author[1*]{Zhenzhong Chen}
\affil[1]{School of Remote Sensing and Information Engineering, Wuhan University}
\affil[2]{School of Data Science, Lingnan University}

\begin{document}

\twocolumn[ % Method A for two-column formatting
  \begin{@twocolumnfalse} % Method A for two-column formatting
  
\maketitle

\begin{abstract}

Structural pruning techniques are essential for deploying multimodal large language models (MLLMs) across various hardware platforms, from edge devices to cloud servers. However, current pruning methods typically determine optimal strategies through iterative search processes, resulting in substantial computational overhead for on-demand MLLMs adaptation. To address this challenge, we propose \textsf{LOP}, an efficient neural pruning framework that learns optimal pruning strategies from the target pruning constraint, eliminating the need for computationally expensive search-based methods. \textsf{LOP} approach trains autoregressive neural networks (NNs) to directly predict layer-wise pruning strategies adaptive to the target pruning constraint, eliminating the time-consuming iterative searches. Experimental results across multiple tasks show that \textsf{LOP} outperforms state-of-the-art pruning methods in various metrics while achieving up to \textit{three orders of magnitude} speedup.

\end{abstract}

\vspace{0.4cm}

  \end{@twocolumnfalse} % Method A for two-column formatting
] % Method A for two-column formatting

\newcommand\blfootnote[1]{%
\begingroup
\renewcommand\thefootnote{}\footnote{#1}%
\addtocounter{footnote}{-1}%
\endgroup
}

\section{INTRODUCTION}
\label{intro.}
{\blfootnote{Corresponding author: Zhenzhong Chen, E-mail:zzchen@ieee.org}}

Multimodal large language models (MLLMs)~\cite{chen2024internvl,yao2024minicpm,lu2024deepseek} have demonstrated impressive capabilities in vision-language tasks, with state-of-the-art models like GPT-4V~\cite{openai2024gpt4technicalreport} and LLaVA~\cite{liu2023visual} achieving over 80\% accuracy on benchmarks such as VQAv2~\cite{antol2015vqa} and TextCaps~\cite{sidorov2020textcaps} for image understanding. However, these models typically require large memory and billions of floating-point operations (FLOPs) per inference, prohibiting deployment across resource-constrained devices. 

Recent advances in model compression have yielded three principal approaches for efficient inference: quantization, distillation, and pruning. While quantization~\cite{frantar2022gptq,lin2024awq} reduces precision and distillation~\cite{gu2024minillm} transfers knowledge to smaller models, pruning offers distinct advantages by eliminating redundant parameters~\cite{frantar2023sparsegpt}, including individual weights, entire neurons, or attention tokens - while preserving the structural model architecture. This selective reduction approach provides exceptional flexibility, with empirical studies demonstrating 40-60\% parameter reduction while maintaining model accuracy, enabling it as an indispensable technique for latency-critical applications.

In MLLMs, the majority of parameters reside in the language backbone; more specifically, within each Transformer layer of that backbone, roughly 73.7\% trainable weights belong to the Feed-Forward Network (FFN), making the FFN the principal driver of overall model size~\cite{wei2024building}. In contrast, the visual encoder and projection layers are relatively lightweight yet more sensitive to perturbations, making them less suitable for aggressive pruning~\cite{liang2025efficientllava,wang2024exploring}. Consequently, current research efforts primarily target FFN weight matrices for pruning. Formally, the pruning challenge constitutes a constrained optimization problem that can be decomposed into two fundamental components:

\begin{enumerate}[0]
\item[$\bullet$] \textbf{Neuron Importance Estimation in MLLMs.} Modern MLLMs often comprise tens to hundreds of billions of trainable parameters, with empirical studies revealing significant redundancy in their weight distributions. Recent analyses demonstrate that plenty of weights in typical transformer-based architectures can be pruned with minimal accuracy loss. Consequently, effective importance estimation methods, ranging from simple magnitude-based criteria to more sophisticated gradient-aware approaches, enable the identification of these non-essential parameters for pruning.
 
\item[$\bullet$]\textbf{Layer-Adaptive Pruning Allocation and Neuron Selection.}  Transformer-based architectures in MLLMs exhibit two critical characteristics that complicate pruning: (1) strong inter-layer dependencies, where early layer outputs condition later layer computations; and (2) non-uniform importance distributions across layers. These observations necessitate intelligent, layer-wise pruning allocation to pursue the better accuracy-efficiency trade-off. Monte Carlo Tree Search~\cite{browne2012survey} is well-suited for this task, as it efficiently explores large, discrete configuration spaces under global constraints and adaptively balances exploration and exploitation. It provides a principled framework for discovering high-quality layer-wise pruning ratios given a target pruning constraint. Given the layer-wise pruning ratio, redundant neurons can be determined based on the neuron importance statistics, i.e., retaining the neurons with high importance. 

\end{enumerate}
\begin{figure*}[t]
  \centering
   \resizebox{1\linewidth}{!}{
  \begin{subfigure}[t]{0.48\linewidth}
    \centering
    \includegraphics[width=\linewidth]{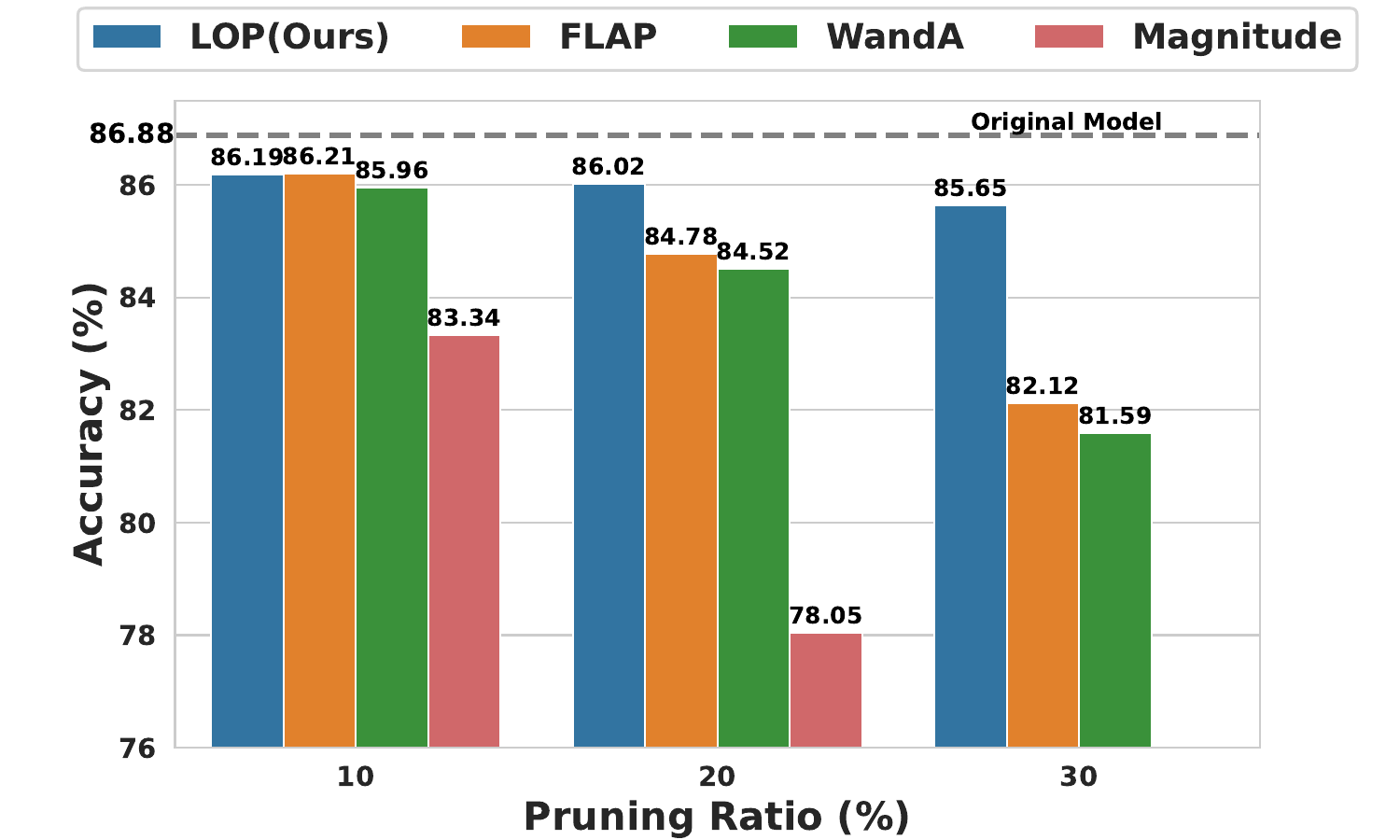}
    \caption{}
    \label{fig:sub1}
  \end{subfigure}
  \hfill
  \begin{subfigure}[t]{0.48\linewidth}
    \centering
    \includegraphics[width=\linewidth]{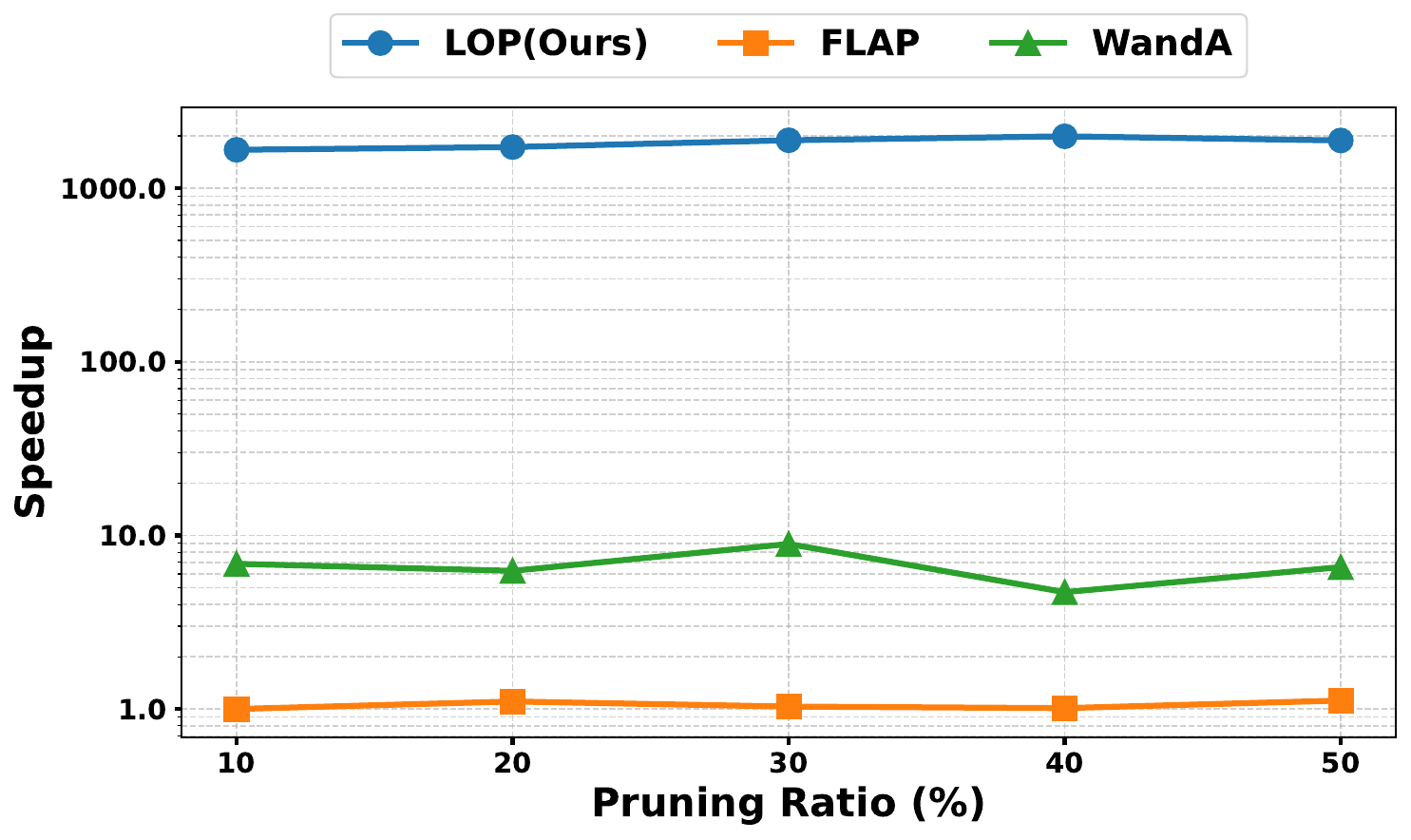}
    \caption{}
    \label{fig:sub2}
  \end{subfigure}
  }
  \caption{(a) \textbf{Accuracy Across Different Pruning Ratios.} A comparison of different methods on MMBench under various pruning ratios shows that \textsf{LOP} maintains strong performance across all sparsity levels. (b) \textbf{Speedup Across Different Pruning Ratios.} Speedup is calculated relative to the runtime of FLAP at a 10\% pruning ratio, with \textsf{LOP} exhibiting notably improved efficiency. Across all sparsity levels, \textsf{LOP} sustains consistently high speedups.}
  \label{fig:combined}
\end{figure*}

Despite the promising performance, search-based optimization approaches face growing computational challenges due to the expansion of possible configurations in modern architectures. To overcome this fundamental limitation, we introduce \textsf{LOP}, a learning-based framework that reformulates pruning strategy discovery as a neural approximation problem. Our idea is to train a NN to learn the implicit mapping between model structure and optimal pruning strategies. Once trained, the network can directly infer near-optimal pruning decisions for any given model and target sparsity level. Compared to manually designed heuristics or exhaustive search, this data-driven approach captures nonlinear dependencies between model architecture and performance in one-shot, significantly reducing computation complexity while maintaining desirable performance across different pruning ratios. In summary, our contributions are three-fold: 
\begin{enumerate}[0]
\item[$\bullet$] We propose \textsf{LOP}, a neural network-based solver for pruning optimization. Specifically, we first design a Monte Carlo Tree Search (MCTS)-based sampling strategy to explore pruning configurations as ground truth using L2-norm-based neuron importance metrics.
% generate training data by estimating neuron importance based on the of activations and adopting 
\item[$\bullet$] We propose an autoregressive modeling approach that makes layer-wise optimal pruning decisions in a sequential manner. This design explicitly captures inter-layer dependencies in pruning ratios and enables fast inference of high-quality pruning configurations across varying sparsity levels.
\item[$\bullet$] Experimental results show that \textsf{LOP} achieves speedups of up to 1567.9$\times$ on multimodal benchmarks while maintaining competitive accuracy with state-of-the-art approaches, and consistently performs well across various multimodal inference tasks. We visualize the performance in Fig. \ref{fig:combined}.

\end{enumerate}

\begin{figure*}[t]
    \centering
    \includegraphics[width=1\linewidth]{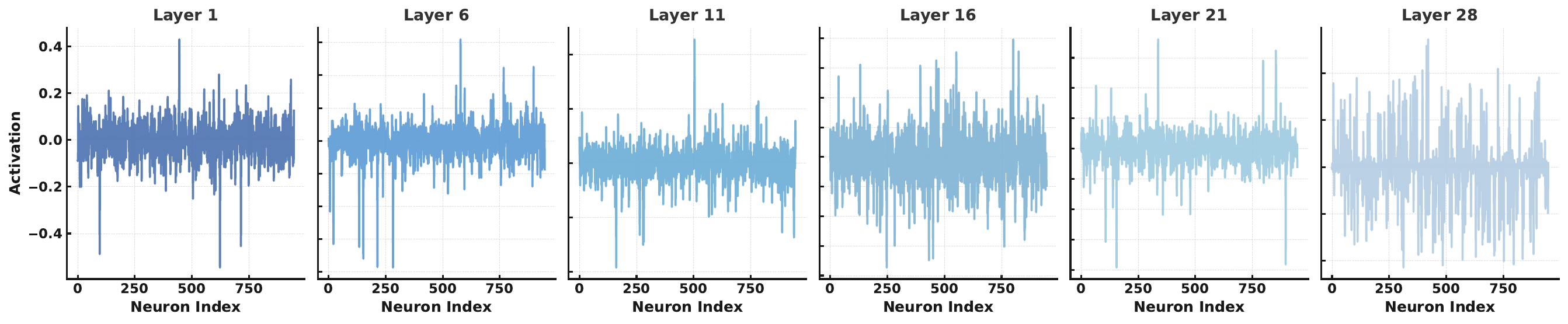}
    \caption{\textbf{Neuron importance distributions and activation curves for six statistically distinct FFN layers.} Each subplot visualizes the activation values of neurons (downsampled for clarity), highlighting distinct patterns of sparsity, intensity, and variation at different depths.}
    \label{neuron.importantce}
\end{figure*}

\section{Related Work}
\label{sec2}
\subsection{LLM Pruning}
The size of Transformer-based large language models (LLMs) has surged to hundreds of billions of parameters. To enable efficient deployment of these massive models across diverse hardware—from edge devices to cloud servers—effective pruning techniques have become indispensable. Current pruning approaches mainly fall into two categories: \textit{static} methods and \textit{dynamic} methods. \textit{static} methods process off-line pruning and permanently remove model components to accelerate all input instance inference. For example, \textsf{WandA}~\cite{wanda} removes less-important weights based on a metric that considers the weights' importance (e.g., activated value). \textsf{SliceGPT}~\cite{ashkboos2024slicegpt} exploits computational invariance in Transformer networks, using principal component analysis (PCA) to remove entire rows and columns of weight matrices. \textsf{LLM-Pruner}~\cite{ma2023llm} is proposed by defining broad sub-structural dependency groups and guiding the pruning process based on significance estimation. Recent work explores pruning within the LLM in MLLM. \textsf{YOPO}~\cite{zhang2024treat} applies pruning strategies at both the parameter and computational pattern levels within MLLMs, achieving significant computation reduction while maintaining competitive performance. In contrast, dynamic methods adapt computations based on input characteristics. For example, \textsf{DejaVu}~\cite{liu2023deja} proposed in adjusts its pruning strategy according to input-specific sparsity patterns, maintaining model accuracy. LLM-\textsf{Streamline}~\cite{chen2024streamlining} identifies less important layers and removes them to reduce model size. This flexibility comes at the cost of increased inference complexity compared to static methods. As discussed in Section~\ref{intro.}, both static and dynamic pruning methods rely on search-based optimization strategies, creating significant computational challenges for modern (and the future) LLMs.
\subsection{Efficient MLLM}

Multimodal large language models (MLLMs) face significant computational overhead and latency when processing images, videos, and other non-text modalities~\cite{jin2024efficient}. This has made efficient inference a critical focus for MLLM research. Existing approaches can be divided into two categories:
\paragraph{Low-complexity Architecture.} Early efforts~\cite{chu2023mobilevlm,zhu2024llava,zhou2024tinyllava,chen2024vitamin} focused on designing lightweight vision encoders or utilizing tiny language models to achieve efficient inference. Mipha~\cite{zhu2024mipha}explores the design space of multimodal small language models by integrating small language models with vision encoders. For example, MoE-LLaVA~\cite{lin2024moe}introduces a sparse mixture-of-experts design into vision-language models, dynamically activating a subset of specialized experts to enhance efficiency. While these approaches improve efficiency and reduce inference cost, they sacrifice generalization ability and require fine-tuning to balance performance and sparsity.
\paragraph{Token Redundancy Reduction.} Recent advances discovered that MLLMs tend to generate excessive visual tokens compared to text-only models. This insight has motivated growing interest in token pruning techniques that streamline input sequences by removing redundant elements. For example, \textsf{FastV}~\cite{chen2024image} identifies important visual tokens using adaptive attention mechanisms in the shallow layers and prunes less critical tokens in deeper layers to enhance computational efficiency. \textsf{SparseVLM}~\cite{zhang2024sparsevlm} prunes redundant visual tokens based on text-visual token relevance, enabling acceleration of vision-language models without retraining. While these methods effectively reduce the computational cost, they do not remove the redundancy in the model's parameters and risk discarding important visual features during inference.

\section{Revisiting Optimal Structual Pruning (OSP) for MLLMs}\label{3}
\subsection{Problem Formulation}
Structural pruning for multimodal large language models aims to reduce the model’s complexity (upon the target pruning constraint) by (i) determining the ratio of neurons to prune for each layer and (ii) selecting which neurons to prune within each layer, while preserving its ability to represent and fuse cross-modal information for comparable performance the downstream task. We formalize this problem as a constrained optimization as follows. 

To begin with, suppose the model contains $L$ transformer layers. Let the Feed-Forward Network (FFN) at (transformer) layer $l$ have $d_l$ neurons. We define a binary selection vector $\mathbf{z}^{(l)}\in \{0,1\}^{d^l}$, where $\mathbf{z}_{i}^{(l)}=1$ indicates the $i$-th neuron is kept, while $\mathbf{z}_{i}^{(l)}=0$ denotes it is pruned. The total number of kept neurons in layer $l$ is $||\mathbf{z}^{(l)}||_{0}$. Let pruning ratio $\mathbf{\theta}^l \in [0,1]$ represent the ratio of pruned neurons in layer $l$. Accordingly, we can have the following constraint for $\mathbf{z}^l$ and $\theta^l$: $\|\mathbf{z}^{(l)}\|_0 = \lfloor (1 - \theta^l) \cdot d^l \rfloor.$

Let $\{x, y\}$ be an input-label pair, where $x \in \mathcal{X},y \in \mathcal{Y}, N \triangleq |\mathcal{X}|$. The Optimal Structural Pruning (OSP) problem can be formulated as:
\begin{equation}
\begin{aligned}
\max_{\{\mathbf{z}^{(l)}\}_{l=1}^L,\{\theta^{(l)}\}_{l=1}^L} \quad & \sum_{x \in \mathcal{X} }c\left(f_{\text{mllm}}(x,\{\mathbf{z}^{(l)}\}_{l=1}^L),y \right) \\
\text{s.t.} \quad & \|\mathbf{z}^{(l)}\|_0 = \lfloor (1 - \theta^l) \cdot d^l \rfloor, \\
                  & \sum_{l=1}^L \theta^l \cdot d^l \leq B_{\text{total}},
\end{aligned}
\label{OSP}
\end{equation}
where $c(\cdot)$ is the performance evaluation metric and $f_{\text{mllm}}(x,\{\mathbf{z}^{(l)}\})$ denotes the output of mllm upon the input $x$ and pruned decision $\{\mathbf{z}^{(l)}\}_{l=1}^L$. $B_{\text{total}}$ is the pruning constraint (i.e., typically representing the maximum allowable number of pruned neurons). We note that the OSP problem is computationally challenging to solve due to (i) the large number of decision variables (e.g., 18944 variables for Qwen2.5-VL-7B per layer, 28 layers in total) and (ii) the dependence between variables. Specifically, the pruning configuration across layers cannot be independently determined as earlier layers in the model influence the representational capacity and input distributions of subsequent layers. 

\subsection{State-of-the-art (SOTA) Pruning Methods}\label{sota.pruning.methods}
SOTA pruning methods typically follow a two-stage process: first, evaluating neuron importance using a carefully designed calibration set, then determining layer-wise pruning ratios and selecting specific neurons to remove. As shown in Fig.~\ref{neuron.importantce}, this approach stems from the empirical finding that neurons exhibit varying degrees of contribution to model performance. The evaluation metric commonly associates higher activation magnitudes with greater importance. Subsequent optimization employs iterative search to determine optimal pruning ratios per layer while preserving top-scoring neurons~\cite{li2025t}. However, these methods face critical limitations—either compromising model performance or demanding excessive computational resources, making them impractical for real-time LLM scaling scenarios. Specifically, the computational bottleneck lies primarily in adaptively determining layer-wise pruning ratios, which typically consumes over 98.4\% of the total processing time. In contrast, both importance evaluation and pruned neuron selection require minimal computational overhead, about 1.6\% of the total processing time. Accordingly, our work addresses the computational challenges in the searching process through a novel learning-based framework that achieves both efficiency and accuracy.

\begin{figure*}[t]
    \centering
    \includegraphics[width=1\linewidth]{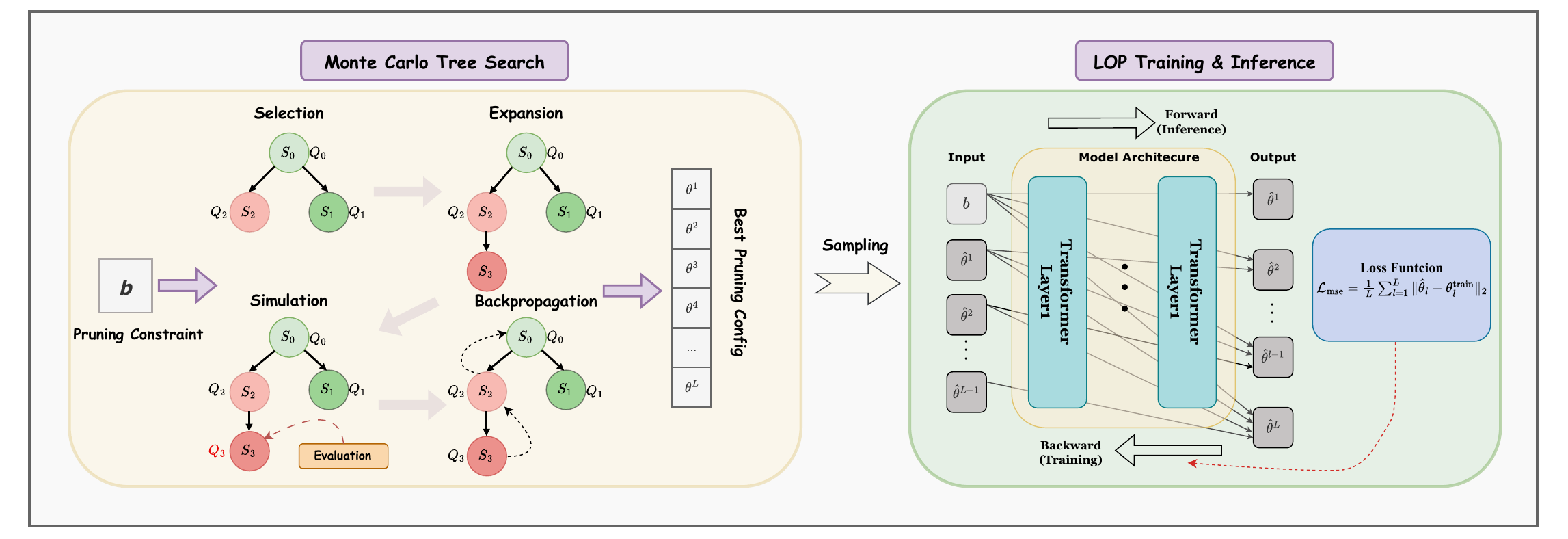}
    \caption{\textbf{Overall Workflow of \textsf{LOP}}. Our approach employs a two-phase framework: first, Monte Carlo Tree Search generates training samples; second, these samples are used to train a Transformer-based autoregressive prediction model.}
    \label{overrall.workflow}
\end{figure*}
\section{LOP: Learning Optimal Pruning}\label{Method}
This section introduces our proposed pruning framework, namely \textsf{LOP}, that learns the optimal pruning from the target pruning constraint. The overall architecture is shown in Fig \ref{overrall.workflow}. The idea behind \textsf{LOP} is to train a DNN to learn the mapping from a target pruning constraint to directly predict layer-wise (optimal) pruning strategies based on target pruning, bypassing the need for time-consuming iterative searches. The following sections detail the architecture and implementation.

\subsection{Overview of LOP}
\label{overview}
As discussed in Section~\ref{sota.pruning.methods}, the main computational bottleneck lies primarily in the searching process for determining layer-wise pruning ratios. To overcome this computational challenge, \textsf{LOP} solves the OSP problem by learning a mapping from a target pruning constraint $B_{\text{total}}$ to the layer-wise pruning ratios $\{\theta^l\}_{l=1}^L$. To effectively capture the complex inter-layer dependencies inherent in deep neural networks, we employ an autoregressive modeling framework that sequentially predicts each pruning ratio $\boldsymbol{\theta}^l$ conditionally on the previous ones: $\{\theta^1, \dots, \theta^{l-1}\}$ as:
\begin{equation}
% \theta_0 = \phi (\mathcal{B}; 0), \quad 
\boldsymbol{\theta}^{l} = \phi (b, \theta^1, \dots, \theta^{l-1}), \quad l = 1,...,L,
% \phi: \rightarrow \{\theta_l\}_{l=1}^L, \quad \theta_l \in [0, 1],
\end{equation}
where $b \in \mathcal{B} \subseteq [0, 1]$ denotes the domain of feasible target pruning constraints. The function $\phi$ maps a given target ratio $b \in \mathcal{B}$ to a sequence of layer-wise pruning ratios $\{\theta^1, \dots, \theta^{l-1}\}$, one for each of the $L$ transformer layers. As such,
the learned mapping $\phi$ captures both the characteristic distribution of pruning ratios across layers and their complex interdependencies.

\subsection{Monte Carlo Tree Search based Sampling}
\label{sec:4.2MCTS}
To construct a reliable dataset for training the mapping function \(\phi\), we adopt a Monte Carlo Tree Search (MCTS) strategy to explore the space of valid pruning configurations. In our formulation, each MCTS \textbf{state} \(s\) corresponds to a complete pruning configuration \(\boldsymbol{\theta} = \{\theta^1, \dots, \theta^L\}\), where each $\theta^l$ denotes the pruning ratio for layer \(l\).
% \(\theta^l \in [0.0, 0.9]\) denotes the \textbf{pruning ratio} for layer \(l\).

At each iteration, MCTS selects a promising configuration \(s\) by maximizing the Upper Confidence Bound (UCB), which balances exploitation and exploration:
\begin{equation}
    a^* = \arg\max_{a \in \mathcal{A}(s)} \left[ Q(s,a) + c \sqrt{\frac{\ln N(s)}{N(s,a)}} \right],
\end{equation}
where \(Q(s,a) = W(s,a) / N(s,a)\) is the average reward of taking action \(a\) from state \(s\), and \(W(s,a)\), \(N(s,a)\) are the cumulative reward and visitation count of edge \((s,a)\), respectively. The constant \(c > 0\) controls the exploration-exploitation trade-off.

The selected configuration is expanded via perturbation: for each layer \(l\), a small noise term is sampled as \(\Delta_l \sim \mathcal{U}(-\delta, \delta)\), with a decaying magnitude \(\delta = 0.1 \cdot 0.9^d\), where \(d\) denotes the tree depth. The resulting candidate becomes \(s' = \{\theta^1 + \Delta_1, \dots, \theta^L + \Delta_L\}\), clipped to \([0.1, 1.0]\), forming a local variant of \(s\).

Each complete configuration \(s'\) is evaluated by pruning the model accordingly and computing its reward as validation accuracy:
\begin{equation}
    V(s') = \frac{1}{N} \sum_{i=1}^N \mathbb{I} \left[ \arg\max f_{\boldsymbol{\theta}}(\boldsymbol{x}_i) = y_i \right].
\end{equation}

To ensure constraint compliance, we enforce that the average pruning ratio satisfies \(\frac{1}{L} \sum_{l=1}^L \theta^l \leq b_{\text{train}}\), where \(b_{\text{train}}\) is a predefined target pruning constraint. If valid, the reward \(r = V(s')\) is propagated backward along the visited path, updating each traversed edge:
\begin{equation}
    N(s,a) \leftarrow N(s,a) + 1, \quad W(s,a) \leftarrow W(s,a) + r.
\end{equation}

After performing \(T\) simulations, the optimal pruning configuration is selected as the leaf node with the highest evaluation reward:
\begin{equation}
    s^* = \arg\max_{s \in \mathcal{S}_{\text{valid}}} V(s),
\end{equation}

where \(\mathcal{S}_{\text{valid}}\) denotes the set of feasible configurations satisfying the pruning constraint. The selected pair \((b, \boldsymbol{\theta}^*)\) serves as one supervised training sample. By repeating this process across multiple constraints, we construct a high-quality dataset for learning the mapping function \(\phi\), which generalizes pruning decisions across sparsity levels and model architectures.

\begin{figure*}[t]
  \centering
  %--------------------- 左：算法 ---------------------%
  \begin{subfigure}[t]{0.48\linewidth}  % ← 给左图 43%
    \centering
    \includegraphics[width=\linewidth]{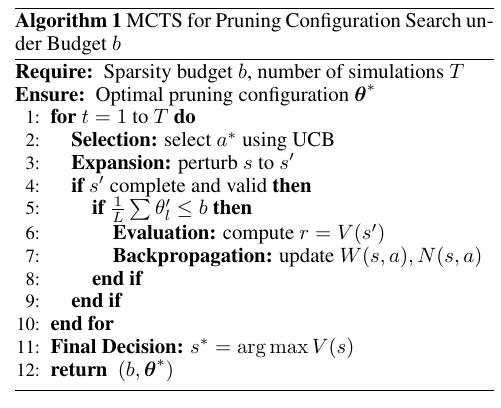}  % 用 width, 不用 height
    %\caption{}  % 如需子图标题可解开
    \label{fig:sub3}
  \end{subfigure}
  \hfill                                   % or \hspace{0.02\linewidth}
  %--------------------- 右：柱状图 -------------------%
  \begin{subfigure}[t]{0.50\linewidth}  % ← 给右图 53%（两者 <100%，预留间隙）
    \centering
    \includegraphics[width=\linewidth]{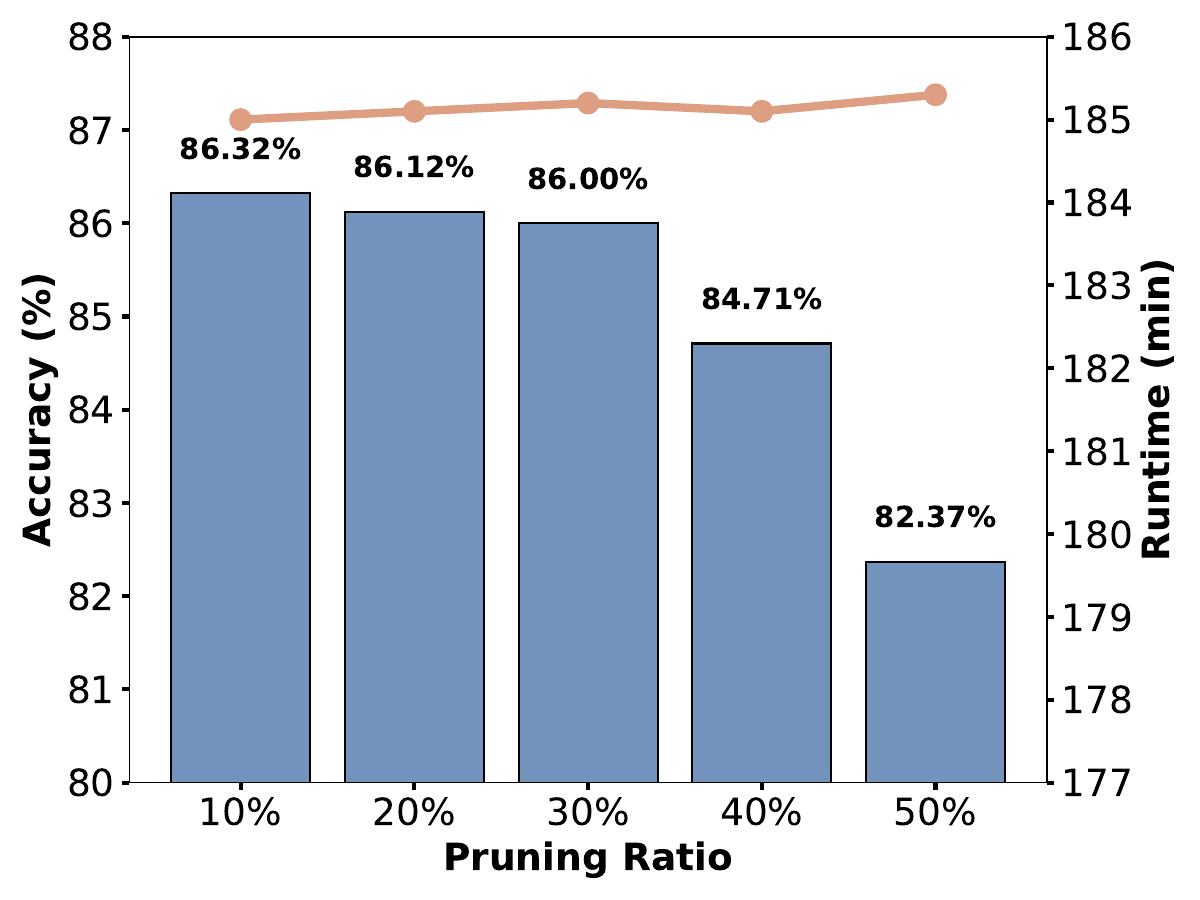}
    %\caption{}
    \label{fig:sub4}
  \end{subfigure}

  \caption{\textbf{Left:} Monte-Carlo Tree Search used to sample pruning configurations under a pruning constraint. 
           \textbf{Right:} Accuracy and runtime at different pruning ratios.  
           Although MCTS yields high-quality configurations, each search still costs $\sim$10 988 s.}
  \label{fig:mcts}
\end{figure*}

As Fig. \ref{fig:mcts} shows, the MCTS procedure outputs the optimal pruning configuration \(\boldsymbol{\theta}^*\) under a target pruning constraint $b$, forming a pair $(b, \boldsymbol{\theta}^*)$ that represents one supervised training sample. By repeating this process with different constraints, we collect a set of high-quality samples ${(b, \boldsymbol{\theta}^{\text{train}})}$ for learning the mapping function $\phi$, which generalizes pruning decisions across varying sparsity levels and model architectures.

\subsection{Design of \textsf{LOP}}
\textbf{Model Architecture.}
We design a Transformer-based autoregressive model to approximate the OSP mapping $\phi(b) \mapsto \{\theta^1, \dots, \theta^L\}$. Unlike parallel regression models, our formulation explicitly models the generation of each pruning ratio $\hat{\theta}_l$ as a conditional prediction based on the target pruning constraint $b$ and all previously predicted ratios $\hat{\theta}_{<l}$:
\begin{equation}
  \hat{\theta}_l = \phi_l(b; \hat{\theta}^1, \dots, \hat{\theta}^{l-1}), \quad l = 1, \dots, L.
\end{equation}
To implement this, we project the  pruning constraint $b$ into a high-dimensional latent vector $\mathbf{x}_0 \in \mathbb{R}^d$ using a two-layer MLP. This vector serves as the conditioning input for all predictions. At step $l$, we construct the input sequence
\begin{equation}
\mathbf{X}^{(l)} = [\mathbf{x}_0,\, \hat{\theta}^1 \mathbf{e}^1,\, \dots,\, \hat{\theta}^{l-1} \mathbf{e}^{l-1}]
\end{equation}
where each $\mathbf{e}_i$ is a learnable embedding for layer $i$. This sequence is fed into a stack of Transformer encoder layers with causal masking, which enforces that each prediction $\hat{\theta}_l$ only attends to $\{\hat{\theta}^1, \dots, \hat{\theta}^{l-1}\}$ and $b$.

Formally, the hidden representation for layer $l$ is computed as:
\begin{align}
\mathbf{h}_{\text{attn}}^{(l)} &= \text{LayerNorm}\left(\mathbf{X}^{(l)} + \text{SelfAttention}(\mathbf{X}^{(l)})\right) \\
\mathbf{h}_l &= \text{LayerNorm}\left(\mathbf{h}_{\text{attn}}^{(l)} + \text{FFN}(\mathbf{h}_{\text{attn}}^{(l)})\right)
\end{align}
and the pruning ratio is predicted as:
\begin{equation}
    \hat{\theta}_l = \sigma(\mathbf{w}^\top \mathbf{h}_l), \quad l = 1, \dots, L
\end{equation}

This autoregressive design enables the model to recursively generate layer-wise pruning decisions, dynamically attending to past decisions and the target pruning constraint. During inference, predictions proceed in a loop: at each step, the model consumes previously predicted $\hat{\theta}_{<l}$ to compute $\hat{\theta}_l$, forming a recurrent policy generation process aligned with the OSP formulation.

\textbf{Loss Function.} All training samples are collected using MCTS-based sampling strategy described in Section~\ref{sec:4.2MCTS}. Each training example generated by MCTS is a pair \((b, \boldsymbol{\theta}^{\text{train}})\), where \(b_{train} \in \mathcal{B}\) is the input target pruning constraint, and \(\boldsymbol{\theta}^{\text{train}} = \{\theta_1^{\text{train}}, \dots, \theta_L^{\text{train}}\}\) is the corresponding high-quality layer-wise pruning ratio sequence sampled. To train the model, we minimize the Mean Squared Error (MSE) between the predicted and ground-truth pruning ratios for each item:
\begin{equation}
    \mathcal{L}_{\text{mse}} = \frac{1}{L} \sum_{l=1}^{L} \Vert\hat{\theta}_l - \theta_l^{\text{train}}\Vert_2
\end{equation}
where \(\theta_l^{\text{gt}}\) denotes the ground-truth pruning ratio for the \(l\)-th layer.

\floatsetup{capposition=top}
\renewcommand{\arraystretch}{1.2} 
\begin{table*}[t]
\caption{Performance Comparisons of Different Pruning Approaches on Qwen2.5-VL-7B.}
\centering
\Large
\resizebox{\textwidth}{!}{%
\begin{tabular}{c|l|cc|cccc|c}
\toprule
\textbf{Ratio} & \textbf{Method} & \textbf{MME-P} & \textbf{MME-R} & \textbf{MMBench} & \textbf{MMMU} & \textbf{POPE} & \textbf{Avg} & \textbf{Speedup} \\
\midrule
0\% & -- & 1694.39 & 640.36 & 86.88 & 51.29 & 87.48 & 75.22 & - \\
\cmidrule{1-9}
\multirow{4}{*}{20\%} 
& Magnitude & 1397.55\dn~\small{(17.5\%)} & 327.86\dn~\small{(48.8\%)} & 78.05\dn~\small{(10.2\%)} & 38.55\dn~\small{(24.8\%)} & 83.76\dn~\small{(4.3\%)} & 66.79\dn~\small{(11.2\%)} & 3034.36 \\
& WandA & 1643.19\dn~\small{(3.0\%)} & 554.64\dn~\small{(13.4\%)} & 84.52\dn~\small{(2.7\%)} & 43.69\dn~\small{(14.8\%)} & \textbf{88.22}\up~\small{(0.8\%)} & 72.14\dn~\small{(4.1\%)} & 5.66 \\
& FLAP & \textbf{1651.26}\dn~\small{(2.5\%)} & 568.57\dn~\small{(11.2\%)} & 84.78\dn~\small{(2.4\%)} & 42.64\dn~\small{(16.9\%)} & 87.67\up~\small{(0.2\%)} & 71.70\dn~\small{(4.7\%)} & 1.00 \\
& \cellcolor{Gray} \textbf{LOP} & \cellcolor{Gray} 1631.15\dn~\small{(3.7\%)} & \cellcolor{Gray}\textbf{605.00}\dn~\small{(5.5\%)} & \cellcolor{Gray} \textbf{86.02}\dn~\small{(1.0\%)} & \cellcolor{Gray} \textbf{46.03}\dn~\small{(10.3\%)} & \cellcolor{Gray} 88.07\up~\small{(0.7\%)} &\cellcolor{Gray} \textbf{73.37}\dn~\small{(2.5\%)} & \cellcolor{Gray} 1560.81 \\
\cmidrule{1-9}
\multirow{4}{*}{30\%}
& Magnitude & 1397.55\dn~\small{(17.5\%)} & 327.86\dn~\small{(48.8\%)} & 58.33\dn~\small{(32.9\%)} & 25.70\dn~\small{(49.9\%)} & 51.41\dn~\small{(41.2\%)} & 45.15\dn~\small{(40.0\%)} & 3246.44 \\
& WandA & 1542.28\dn~\small{(9.0\%)} & 429.64\dn~\small{(32.9\%)} & 81.59\dn~\small{(6.1\%)} & 43.11\dn~\small{(15.9\%)} & \textbf{88.38}\up~\small{(1.0\%)} & 71.03\dn~\small{(5.6\%)} & 7.90 \\
& FLAP & 1537.39\dn~\small{(9.3\%)} & 534.64\dn~\small{(16.5\%)} & 82.12\dn~\small{(5.5\%)} & 41.24\dn~\small{(19.6\%)} & 87.46\dn~\small{(0.0\%)} & 70.27\dn~\small{(6.6\%)} & 1.00 \\
& \cellcolor{Gray} \textbf{LOP} & \cellcolor{Gray}\textbf{1613.73}\dn~\small{(4.8\%)} & \cellcolor{Gray} \textbf{575.00}\dn~\small{(10.2\%)} & \cellcolor{Gray} \textbf{85.65}\dn~\small{(1.4\%)} &\cellcolor{Gray} \textbf{44.16}\dn~\small{(13.9\%)} & \cellcolor{Gray} 87.83\up~\small{(0.4\%)} & \cellcolor{Gray} \textbf{72.55}\dn~\small{(3.5\%)} & \cellcolor{Gray} 1460.90 \\
\cmidrule{1-9}
\multirow{3}{*}{50\%}
& WandA & 1142.19\dn~\small{(32.6\%)} & 132.86\dn~\small{(79.3\%)} & 58.67\dn~\small{(32.5\%)} & 29.09\dn~\small{(43.3\%)} & 85.91\dn~\small{(1.8\%)} & 57.89\dn~\small{(23.0\%)} & 5.87 \\
& FLAP & 1337.49\dn~\small{(21.1\%)} & 356.07\dn~\small{(44.4\%)} & 70.43\dn~\small{(18.9\%)} & 32.83\dn~\small{(36.0\%)} & 87.46\dn~\small{(0.0\%)} & 63.57\dn~\small{(15.5\%)} & 1.00 \\
 & \cellcolor{Gray} \textbf{LOP} & \cellcolor{Gray} \textbf{1478.07}\dn~\small{(12.8\%)} & \cellcolor{Gray} \textbf{468.57}\dn~\small{(26.8\%)} & \cellcolor{Gray} \textbf{82.17}\dn~\small{(5.4\%)} & \cellcolor{Gray} \textbf{35.51}\dn~\small{(30.8\%)} & \cellcolor{Gray} \textbf{87.47}\dn~\small{(0.0\%)} &\cellcolor{Gray} \textbf{68.38}\dn~\small{(9.1\%)} &\cellcolor{Gray} \textbf{1682.39} \\
\bottomrule
\multicolumn{9}{l}{\textit{Note:} Magnitude pruning at 50\% leads to severe degradation, with the model nearly losing its language generation capability.} \\
\end{tabular}%
}
\label{Performance Comparisons}
\end{table*}

\section{Experiments}
\label{others}

\subsection{Experimental Setups}

\paragraph{Multimodal Large Language Model.}%To assess the effectiveness and generalizability of \textsf{LOP}, 
Experiments are conducted on the state-of-the-art multimodal large language models: Qwen2.5-VL-7B~\cite{bai2025qwen2}, which incorporates the Qwen2.5-7B\cite{yang2024qwen2} as the lare language model backbone. 

\paragraph{Baselines.} The performance of \textsf{LOP} is compared with that of three state-of-the-art heuristic pruning approaches: (i) \textsf{Magnitude} pruning, which removes weights with the smallest absolute values, assumes that low-magnitude parameters have minimal impact; (ii) \textsf{WandA}~\cite{wanda} pruning, which removes unimportant weights based on a metric that considers the weights' importance; (iii) \textsf{FLAP}~\cite{an2024fluctuation} pruning, which extends \textsf{WandA} by pruning based on the recoverability of feature outputs after weight removal.
% ;(IV)MonteCarlo

\paragraph{Evaluation Metrics and Datasets.}%To assess the pruned models’ performance on general-purpose tasks, 
The method is evaluated on standard multimodal benchmarks including MMBench~\cite{liu2024mmbench}, MME~\cite{fu2024mmecomprehensiveevaluationbenchmark}, and MMMU~\cite{yue2024mmmu}. MME is adopted for its stringent accuracy+ metric, which assigns credit only when all associated questions for an image are answered correctly. This differs from standard accuracy metrics adopted by other benchmarks, which assign partial credit for individual correct responses.
% MME uses accuracy+ as the evaluation metric, which requires both associated questions for each image to be answered correctly. Other benchmarks use standard accuracy, where higher scores indicate better performance. 
In addition, to examine whether pruning introduces increased hallucination, we further include evaluation on the hallucination detection benchmark, POPE~\cite{2023pope}. We evaluate the running time and speedup of pruning approaches, where the speedup ratio is defined as the average runtime of the FLAP approach divided by that of the test approaches. 

\paragraph{Implementation Details.} For calibration purposes, we randomly select 500 samples from MMBench and compute neuron importance scores as the average L2 norm of activations across this subset. The training data generation employs Monte Carlo sampling with a fixed ratio-step of 0.025. \textsf{LOP}'s adopts NN that consists of 28 hidden layers, each containing 18,944 neurons. We set the batch size, maximum number of epochs, and learning rate to 40, 64, and 1e-3, respectively.

% During pruning, we select 500 samples from MMBench as calibration data. Neuron importance is estimated using the average L2 norm of activations across these samples. Monte Carlo sampling is then applied with varying step sizes to generate training data for the pruning predictor. 

\subsection{Experimental Results}
% \com{Accuracy down,compare to SOTA }
% \textbf{Performance Comparison.}
\subsubsection{Comparison with State-of-the-art Methods}
We compare our method LOP with several representative pruning baselines in terms of pruning performance and efficiency in Table~\ref{Performance Comparisons}. Using FLAP as a reference, we compute the relative speedup of each method. Magnitude-based pruning, which removes weights with the smallest absolute values, is extremely fast due to its simplicity, but leads to drastic performance degradation—in our tests, language capabilities were almost entirely lost, illustrating the limitation of heuristics that rely solely on parameter magnitude. WandA and FLAP maintain decent performance under low pruning ratios but suffer significant degradation at higher pruning levels. In contrast, LOP consistently outperforms all baselines across different pruning ratios, demonstrating its ability to preserve model capacity and performance. Notably, WandA suffers a significant performance drop of 25.15\% on average at a 50\% pruning ratio. In contrast, our method shows a much smaller degradation of only \textbf{11.33\%} under the same pruning level. Moreover, it achieves an average \textbf{1567.9× speedup} compared to FLAP, verifying its practical efficiency for large-scale model compression.

% \textbf{Detailed Analysis.}
% \subsubsection{Performance Degradation after Pruning}
As shown in Table~\ref{Performance Comparisons}, LOP retains strong multimodal capabilities on Qwen2.5-VL under moderate pruning ratios. Models pruned at 20\% and 30\% reach 95.57\% and 94.95\% of the original performance on MMBench and MMMU, respectively, indicating that substantial parameter reduction can be achieved with minimal performance loss.

On the hallucination benchmark POPE, pruning moderately reduces hallucination errors. This improvement may stem from the elimination of redundant parameters, which lowers model complexity and overfitting, resulting in more cautious and reliable generation. However, at 50\% pruning ratio, performance begins to decline, suggesting that overly aggressive pruning harms both model capacity and optimization stability.

Furthermore, analysis on the MME dataset reveals that pruning disproportionately affects different functional categories. As the pruning ratio increases, performance on cognition-related tasks degrades more severely than on perception tasks. At 30\% pruning ratio, accuracy on perception tasks drops by 4.67\%, while cognition tasks see a 10.2\% drop, indicating that reasoning and inference capabilities are more sensitive to parameter reduction.

These findings highlight that while moderate pruning preserves overall multimodal capabilities, excessive sparsity leads to notable degradation—especially in cognition-heavy tasks—underscoring the need for balanced pruning strategies that account for functional sensitivity. Future efforts should also explore ways to bolster the reasoning and cognitive faculties of pruned models so that they remain fully competitive with their dense counterparts.

\subsubsection{Comparison with Other Scaling Methods}
We evaluate our pruned models against smaller-scale multimodal architectures or mixture-of-experts models (with comparable parameter counts), as detailed in Table~\ref{tab:small}.
This comparison demonstrates the effectiveness of pruning versus other scaling methods. 
Notably, the 60\% pruned version outperforms both Mipha-3B ~\cite{zhu2024mipha} and the MoE-based Moe-LLaVA-2.7B$\times4$~\cite{lin2024moe} on MMBench, despite having significantly fewer effective parameters. Similarly, the 50\% pruned model achieves better performance than Intern2.5-VL-4B\cite{chen2024expanding}. These results suggest that pruning large models into lightweight versions can yield superior performance compared to training small models from scratch, particularly in resource-constrained edge environments.

\subsection{Ablation Study}

\textbf{Effects of Different Model Architecture.}
We explore other variants of backbone architectures for predicting layer-wise retention ratios on the MMBench benchmark. As shown in Table~\ref{tab:backbone_ablation}, the Transformer consistently outperforms both Bi-LSTM and MLP across all global pruning ratio settings (50\%, 30\%, and 20\%). In particular, it achieves the highest average accuracy across all pruning ratios, with an average accuracy of \textbf{84.61}. This indicates that the Transformer is better at capturing global-to-layer dependencies for adaptive pruning.

While MLP and Bi-LSTM provide competitive results, the Transformer consistently achieves superior performance, underscoring the strength of its autoregressive modeling. By sequentially generating layer-wise pruning ratios conditioned on previous decisions, it effectively captures inter-layer dependencies and produces globally coherent pruning policies—key to preserving model performance under high pruning ratios.
\floatsetup[table]{capposition=top}
\newfloatcommand{capbtabbox}{table}[][\FBwidth]

\begin{table}[t]
    \centering
    \begin{tabular}{l|ccc}
    \toprule
    \textbf{Model} & \textbf{MMbench} & \textbf{MME}  \\
    \midrule
    Mipha-3B   & 69.7 & 1783.9 \\
    InternVL2.5-4B   & 81.6 & 1939.2  \\
    MoE-LLaVA-2.7B\(\times\)4   & 68.0 & 1431.3 \\
    \midrule
    \textbf{LOP(60\%)}   & 80.0 & 1773.1  \\
    \textbf{LOP(50\%)}   & \textbf{82.2} & \textbf{1946.6}\\
    
    \bottomrule
    \end{tabular}
     \caption{Performance comparison between LOP and other compact multimodal models.}
     \label{tab:small}
     \footnotesize
\end{table}

\begin{table}[t]
    \centering
    \begin{tabular}{l|ccc}
    \toprule
    \textbf{Ratio} & \textbf{Transformer} & \textbf{Bi-LSTM} & \textbf{MLP} \\
    \midrule
    50\%   & \textbf{82.17} & 81.68 & 82.10 \\
    30\%   & \textbf{85.65} & 85.54 & 85.59 \\
    20\%   & \textbf{86.02} & 85.79 & 85.84 \\
    \midrule
    \textbf{Avg} & \textbf{84.61} & 84.34 & 84.51 \\
    \bottomrule
    \end{tabular}
     \caption{MMBench under different global retention ratios using various backbone predictors.}
     \label{tab:backbone_ablation}
     \footnotesize
\end{table}

\section{Conclusions and Future Work}
This work introduces \textsf{LOP}, a neural pruning framework designed specifically for large multimodal models. The framework employs Monte Carlo sampling to systematically gather training data containing optimal pruning configurations. We train a neural network to establish the relationship between global pruning ratios and corresponding layer-wise retention parameters. Through an autoregressive generation approach, LOP efficiently determines appropriate pruning configurations for any desired sparsity level. Comprehensive evaluations across diverse benchmarks show that our framework maintains model accuracy while dramatically reducing pruning computation time by up to \textit{1000$\times$}. The computational efficiency and adaptive nature make it particularly valuable for edge computing applications and dynamic model compression scenarios.

Existing techniques experience pronounced performance degradation once the pruning ratio surpasses 50\%. Our future work is to overcome this bottleneck and preserve multimodal accuracy under even more aggressive pruning. Specifically, we will (i) integrate knowledge-distillation signals during pruning to offset capacity loss, and (ii) jointly optimize pruning with low-bit quantization to fully exploit their complementary compression benefits.
\bibliography{ms}
% that's all folks

\appendix
\clearpage

\section{Appendix}

\subsection{Implementation Details}

\subsubsection{Detailed Network Formulation}
Given a target pruning constraint \( b \in \mathbb{R} \), our goal is to predict a 28-dimensional vector \( \boldsymbol{\theta} \in (0, 1)^{28} \), where each element represents the pruning ratio for a specific transformer layer. We evaluate three model architectures: MLP, Bi-LSTM, and Transformer. Below we formalize each.

\paragraph{MLP Predictor.}
The MLP treats the prediction as a static regression task. It consists of two hidden layers with ReLU activations and a final Sigmoid output:
\begin{align}
\mathbf{h}_1 &= \phi(\mathbf{W}_1 b + \mathbf{b}_1), \\
\mathbf{h}_2 &= \phi(\mathbf{W}_2 \mathbf{h}_1 + \mathbf{b}_2), \\
\boldsymbol{\theta} &= \sigma(\mathbf{W}_3 \mathbf{h}_2 + \mathbf{b}_3),
\end{align}
where \( \phi(\cdot) \) is the ReLU activation and \( \sigma(\cdot) \) is the element-wise Sigmoid function. The architecture assumes that each layer’s pruning ratio is conditionally independent given \( b \).

\paragraph{Bi-LSTM Predictor.}
To model layer-wise dependencies, we project \( b \) into a hidden vector and replicate it across 28 positions, which are then processed by a bidirectional LSTM:
\begin{align}
\mathbf{z} &= f_{\text{proj}}(b), \\
\mathbf{X} &= [\mathbf{z}, \dots, \mathbf{z}] \in \mathbb{R}^{28 \times d}, \\
\mathbf{H} &= \text{BiLSTM}(\mathbf{X}), \\
\boldsymbol{\theta} &= \sigma(\mathbf{W}_{\text{out}} \mathbf{H} + \mathbf{b}_{\text{out}}),
\end{align}
This formulation allows each layer’s decision to be influenced by its neighbors, leveraging both forward and backward temporal context.

\paragraph{Transformer Predictor.}
We further explore a Transformer-based predictor to capture long-range interactions across layers. The global input \( b \) is embedded and combined with learnable positional encodings:
\begin{align}
\mathbf{z} &= f_{\text{embed}}(b) \in \mathbb{R}^{d}, \\
\mathbf{X}_0 &= [\mathbf{z} + \mathbf{p}_i]_{i=1}^{28}, \\
\mathbf{X}_L &= \text{Transformer}(\mathbf{X}_0), \\
\boldsymbol{\theta} &= \sigma(\mathbf{W}_{\text{out}} \mathbf{X}_L + \mathbf{b}_{\text{out}}),
\end{align}
where \( \mathbf{p}_i \in \mathbb{R}^{d} \) are trainable positional embeddings. The Transformer encoder allows for modeling complex inter-layer dependencies via multi-head self-attention.

\subsubsection{LOP Predictor Configuration.}

The Transformer encoder used for layer-wise pruning prediction is configured as follows:
\begin{itemize}
  \item \textbf{Number of layers:} 2
  \item \textbf{Hidden dimension (\(d\)):} 128
  \item \textbf{Number of attention heads:} 4
  \item \textbf{Feed-forward expansion:} 4×
  \item \textbf{Activation function:} GELU
  \item \textbf{Sequence length:} 28 (one per transformer layer)
  \item \textbf{Positional encoding:} Learnable
  \item \textbf{Output layer:} Linear + Sigmoid
\end{itemize}

We performed all experiments using NVIDIA RTX 3090 and A100 \(\times\) 40G.

\subsection{Neuron Importance Evaluation}
\label{sec:importance}

We primarily assess neuron importance based on the L2 norm of their activation values. In Transformer-based multimodal large language models, Feed-Forward Networks (FFNs) play a critical role by enhancing the model's ability to capture nonlinear interactions between input features. Each FFN layer consists of two linear projections interconnected by a nonlinear activation function, formulated as follows:

\begin{equation} 
\mathbf{h}_l = \sigma\left( \mathbf{W}_1^{(l)} \mathbf{x} + \mathbf{b}_1^{(l)} \right), \quad \mathbf{y}_l = \mathbf{W}_2^{(l)} \mathbf{h}_l + \mathbf{b}_2^{(l)} 
\end{equation}

where \(\sigma\left(·\right)\) denotes the nonlinear activation function,\(W_{1}^{l}\),\(W_2^{l}\) are weight matrices, and  \(b_{1}^{l}\),  \(b_{2}^{l}\)  are corresponding bias vectors.

The core idea behind neuron pruning lies in the observation that individual neurons differ significantly in their contribution to the model’s representational capacity. Specifically, neurons that exhibit consistently higher activation magnitudes tend to encode more significant semantic or structural information crucial to model performance. Thus, we propose a neuron importance metric derived from the L2 norm of activations across a representative dataset, defined formally as:
\begin{equation} 
\text{Importance}j = \sqrt{ \frac{1}{N} \sum_{i=1}^N \left[ \phi\left(\boldsymbol{w}_j^\top \boldsymbol{x}^{(i)}\right) \right]^2 } 
\end{equation}
where  \(\sigma\left(·\right)\) is the activation function, \(w_j\) represents weights corresponding to the j-th neuron, \(x^{i}\)  is the input vector from the \(i\)-th data sample, and \(N\) is the total number of samples utilized for estimating neuron importance. Neurons with higher importance scores are retained preferentially during pruning, whereas those with lower scores are removed to reduce computational complexity while minimally impacting the overall performance.

\subsection{Procedure Monte Carlo Tree Search}
We adopt Monte Carlo Tree Search (MCTS) to explore layer-wise pruning configurations under a global constraint. The search process is conducted on the MMBench validation set, using validation accuracy as the reward signal. We perform 300 simulations per search, evaluate 200 sampled configurations, and use a decaying perturbation step with initial magnitude \(\delta =0.1\). 
The procedure consists of four stages: selection, expansion, evaluation, and backpropagation, as detailed below.

\textbf{Step1 - Selection.}
In the selection phase, we utilize the Upper Confidence Bound (UCB) for Trees to balance exploration and exploitation, guiding the search towards promising configurations while maintaining sufficient exploration of less-visited regions. Formally, the selection criterion is:
\begin{equation} 
a^* = \mathop{\arg\max}_{a \in \mathcal{A}(s)} \left[ Q(s,a) + c \sqrt{\frac{\ln N(s)}{N(s,a)}} \right]
\end{equation}
Here, $Q(s,a) = W(s,a) / N(s,a)$ denotes the \textbf{average reward} of taking action $a$ from state $s$, where $W(s,a)$ and $N(s,a)$ are the cumulative reward and visitation count of edge $(s,a)$, respectively. $N(s)$ is the total visitation count for state $s$, and $c > 0$ is a constant controlling the exploration-exploitation trade-off.

\textbf{Step2 - Expansion.}
When the selected state $s = \langle \theta_1, \dots, \theta_l \rangle$ has not been fully expanded, we generate new candidate configurations by perturbing all pruning ratios assigned so far. For each $k = 1, \dots, l$, we sample:
\begin{align}
\Delta_k &\sim \mathcal{U}(-\delta, \delta),\quad \delta = 0.1 \times 0.9^{d} \\
\theta_k' &= \text{clip}(\theta_k + \Delta_k,\ 0.1,\ 1.0)
\end{align}
Here, $d$ is the current depth of the search tree, and $\delta$ is a decaying perturbation range that ensures coarse-to-fine granularity. The resulting child node becomes $s' = \langle \theta_1', \dots, \theta_l' \rangle$, representing a local variant of the current pruning decision path.

\textbf{Step3 - Configuration Evaluation.}
Once a complete configuration $s_{\text{leaf}} = \langle \theta_1, \dots, \theta_L \rangle$ is reached, the corresponding pruned model $f_{\boldsymbol{\theta}}$ is evaluated on a held-out validation set. The validation accuracy is used as the reward:
\begin{equation}
V(s_\text{leaf}) = \frac{1}{N} \sum_{i=1}^N \mathbb{I} \left[ \mathop{\text{argmax}} \left( f_{\boldsymbol{\theta}}(\boldsymbol{x}_i) \right) = y_i \right]
\end{equation}

\textbf{Step4 - Reward Propagation.}
The reward $r = V(s_\text{leaf})$ is propagated back along the path, updating each visited edge $(s,a)$ as:
\begin{align}
N(s,a) &\leftarrow N(s,a) + 1 \\
W(s,a) &\leftarrow W(s,a) + r
\end{align}

\textbf{Final Selection.}
After $T$ simulations, the optimal pruning configuration is selected as the leaf node with the highest evaluation reward:
\begin{equation}
s^* = \arg\max_{s \in \mathcal{S}_{\text{valid}}} V(s)
\end{equation}
where $\mathcal{S}_{\text{valid}}$ denotes the set of complete configurations satisfying the global pruning constraint $\frac{1}{L} \sum_{l=1}^L \theta_l \leq b_{\text{train}}$.

% \subsection{Experimental Results on InternVL2.5-8B}

% \subsection{Ablation of the Number of Transformer Layers in LOP}

\subsection{Limitations and Future Work}
While our method demonstrates strong performance under moderate sparsity, we observe a significant accuracy drop when the pruning ratio exceeds 50\%. This highlights a key limitation in maintaining multimodal alignment under extreme compression. Moreover, our current approach primarily prunes the feed-forward networks (FFNs) in multimodal large models, while the attention layers remain largely unpruned. Prior studies have shown that different attention heads vary in importance, suggesting that head-wise pruning could further enhance compression efficiency. Future work will address these limitations by (i) incorporating knowledge distillation to mitigate capacity loss, (ii) extending pruning to the attention mechanism, and (iii) jointly optimizing pruning with low-bit quantization to exploit their complementary advantages.

\end{document}